%% file: main.tex
\documentclass[nohyperref]{article}

\input{math_commands.tex}
\usepackage{microtype}
\usepackage{graphicx}
\usepackage{subfigure}
\usepackage{booktabs} %

\usepackage{hyperref}

\usepackage[accepted]{icml2022}

\usepackage{amsmath}
\usepackage{amssymb}
\usepackage{mathtools}
\usepackage{amsthm}

\usepackage[capitalize,noabbrev]{cleveref}

\theoremstyle{plain}
\newtheorem{thm}{Theorem}[section]
\newtheorem{prop}[thm]{Proposition}
\newtheorem{lem}[thm]{Lemma}

\theoremstyle{definition}
\newtheorem{defn}[thm]{Definition}

\theoremstyle{remark}
\newtheorem{rmk}[thm]{Remark}
\newtheorem{property}[thm]{Property}

\usepackage{wrapfig}
\usepackage{threeparttable}
\usepackage{multirow}

\ifodd 0
\newcommand{\rev}[1]{{\color{blue}#1}}
\newcommand{\yl}[1]{\textbf{\color{red}(Yang: #1)}}
\newcommand{\zzw}[2]{\textbf{\color{blue}(Zhaowei: #1)}{\color{blue}~#2}}

\newcommand{\clar}[1]{\textbf{\color{green}(NEED CLARIFICATION: #1)}}

\else
\newcommand{\rev}[1]{#1}
\newcommand{\com}[1]{}
\newcommand{\clar}[1]{}
\newcommand{\response}[1]{}
\newcommand{\yl}[1]{}
\newcommand{\zzw}[2]{}

\fi

\newcommand{\algcom}[1]{\textsl{\color{blue}{\footnotesize #1}}}

\newcommand{\ours}{\textsf{SimiFeat}}

\definecolor{myorange}{RGB}{241,167,108}
\definecolor{myblue}{RGB}{94,167,243}
\definecolor{mygreen}{RGB}{92,199,60}

\usepackage[textsize=tiny]{todonotes}

\icmltitlerunning{Detecting Corrupted Labels Without Training a Model to Predict}

\begin{document}

\twocolumn[
\icmltitle{Detecting Corrupted Labels Without Training a Model to Predict}

\icmlsetsymbol{equal}{*}
\begin{icmlauthorlist}
\icmlauthor{Zhaowei Zhu}{UCSC}
\icmlauthor{Zihao Dong}{UCSC}
\icmlauthor{Yang Liu}{UCSC}
\end{icmlauthorlist}

\icmlaffiliation{UCSC}{Department of Computer Science and Engineering, University of California, Santa Cruz, CA, USA}

\icmlcorrespondingauthor{Yang Liu}{yangliu@ucsc.edu}

\icmlkeywords{Machine Learning, ICML}

\vskip 0.3in
]

\printAffiliationsAndNotice{}  %

\begin{abstract}
Label noise in real-world datasets encodes wrong correlation patterns and impairs the generalization of deep neural networks (DNNs). It is critical to find efficient ways to detect corrupted patterns. Current methods primarily focus on designing robust training techniques to prevent DNNs from memorizing corrupted patterns. These approaches often require customized training processes and may overfit corrupted patterns, leading to a performance drop in detection. In this paper, from a more data-centric perspective, we propose a training-free solution to detect corrupted labels. Intuitively, ``closer'' instances are more likely to share the same clean label. Based on the neighborhood information, we propose two methods: the first one uses ``local voting" via checking the noisy label consensuses of nearby features. The second one is a ranking-based approach that scores each instance and filters out a guaranteed number of instances that are likely to be corrupted. We theoretically analyze how the quality of features affects the local voting and provide guidelines for tuning neighborhood size. We also prove the worst-case error bound for the ranking-based method. Experiments with both synthetic and real-world label noise demonstrate our training-free solutions consistently and significantly improve most of the training-based baselines. Code is available at \url{github.com/UCSC-REAL/SimiFeat}.
\end{abstract}

\input{src_icml/intro}

\input{src_icml/related}

\input{src_icml/method}

\input{src_icml/analyses}

\input{src_icml/exp}

\input{src_icml/conclusion}

\clearpage
\newpage
\bibliography{ref}
\bibliographystyle{icml2022}

\newpage
\appendix
\onecolumn
\input{src_icml/appendix}

\end{document}

%% file: math_commands.tex
\usepackage{amsmath,amsfonts,bm,dsfont,amsthm}

\def\eqref#1{equation~\ref{#1}}

\def\1{\bm{1}}

\def\ve{{\bm{e}}}

\def\vy{{\bm{y}}}

\DeclareMathAlphabet{\mathsfit}{\encodingdefault}{\sfdefault}{m}{sl}
\SetMathAlphabet{\mathsfit}{bold}{\encodingdefault}{\sfdefault}{bx}{n}

\DeclareMathOperator*{\argmax}{arg\,max}

\newcommand{\PP}{\mathbb P}

\newcommand{\BR}{\mathds 1}

\usepackage{tcolorbox}
\definecolor{grey}{rgb}{0.33, 0.33, 0.33}

\newcommand{\squishlist}{
\begin{list}{{{\small{$\bullet$}}}}
{\setlength{\itemsep}{1pt}      \setlength{\parsep}{0pt}
\setlength{\topsep}{-2pt}       \setlength{\partopsep}{0pt}
\setlength{\leftmargin}{1em} \setlength{\labelwidth}{1em}
\setlength{\labelsep}{0.5em} } }
\newcommand{\squishend}{  \end{list}  }

\makeatletter
\renewcommand*\env@matrix[1][*\c@MaxMatrixCols c]{%
  \hskip -\arraycolsep
  \let\@ifnextchar\new@ifnextchar
  \array{#1}}
\makeatother

%% file: src_icml/intro.tex
\section{Introduction}\label{sec:intro}

The generalization of deep neural networks (DNNs) depends on the quality and the quantity of the data. Nonetheless, real-world datasets often contain label noise that challenges the above assumption \citep{krizhevsky2012imagenet,zhang2017improving,agarwal2016learning,wang2021policy}.
Employing human workers to clean annotations is one reliable way to improve the label quality, but it is too expensive and time-consuming for a large-scale dataset. One promising way to automatically clean up label errors is to first algorithmically detect possible label errors from a large-scale dataset \citep{cheng2021learningsieve,northcutt2021confident,pruthi2020estimating,bahri2020deep}, and then correct them using either algorithm or crowdsourcing \citep{northcutt2021pervasive}.

Almost all the algorithmic detection approaches focus on designing customized training processes to learn with noisy labels, where the idea is to train DNNs with noisy supervisions and then make decisions based on the output \citep{northcutt2021confident} or gradients \citep{pruthi2020estimating} of the last logit layer of the trained model. The high-level intuition of these methods is the memorization effects \citep{han2020survey}, i.e., instances with label errors, a.k.a., corrupted instances, tend to be harder to be learned by DNNs than clean instances \citep{xia2021robust,liu2020early,bai2021me}.
By setting appropriate hyperparameters to utilize the memorization effect, corrupted instances could be identified.

\rev{\textbf{Limitations of the learning-centric methods~}}
The above methods suffer from \emph{two major limitations}: 1) the customized training processes are task-specific and may require fine-tuning hyperparameters for different datasets/noise; 2) as long as the model is trained with noisy supervisions, the memorization of corrupted instances exists. The model will ``subjectively'' and wrongly treat the memorized/overfitted corrupted instances as clean.  For example, some low-frequency/rare clean instances may be harder to memorize than high-frequency/common corrupted instances.
Memorizing these corrupted instances lead to unexpected and disparate impacts \citep{liu2021importance}.

\begin{figure}[t]
    \centering
    \includegraphics[width=0.48\textwidth]{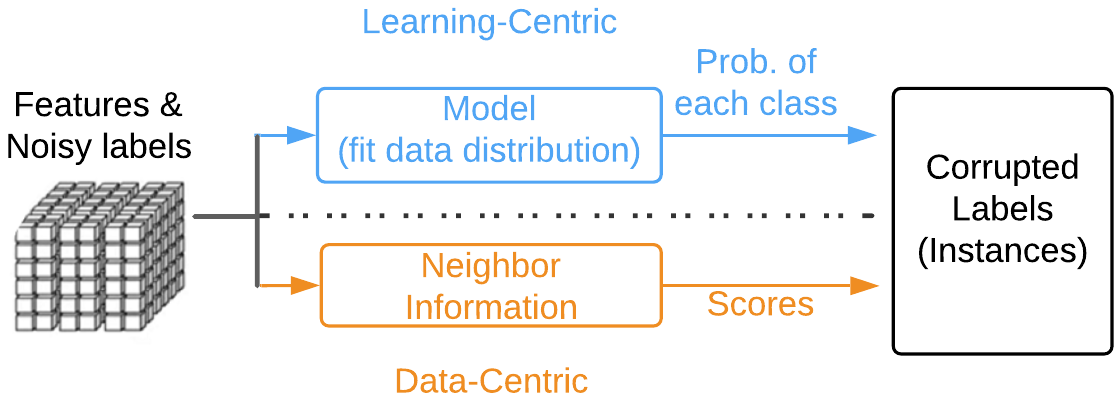}
    \vspace{-12pt}
    \caption{The existing learning-centric pipeline vs. our proposed data-centric pipeline. The inputs are features and the corresponding noisy labels, and the outputs are a set of corrupted labels. {\color{myblue}Blue:} The learning-centric solution. {\color{myorange}Orange:} The data-centric solution.}
    \label{fig:data-centric}
    \vskip -0.11in
\end{figure}

\rev{Existing solutions to avoid memorizing/overfitting are to employ some regularizers \cite{cheng2021learningsieve} or use early-stopping \cite{bai2021understanding,li2020gradient}. However, their performance depends on hyperparameter settings.
One promising way} to avoid memorizing/overfitting is to drop the dependency on the noisy supervision, which motivates us to design a \emph{training-free} method to find label errors.
Intuitively, we can carefully use the information from nearby features to infer whether one instance is corrupted or not. The comparison between our data-centric and existing learning-centric solutions is illustrated in Figure~\ref{fig:data-centric}.

Our training-free method enables more possibilities beyond a better detection result.
For example, the concerns about the required assumptions and hyperparameter tuning in those training-based methods will now be released due to our training-free property.
The complexity will also be much lower, again due to removing the possibly involved training processes. This light detection solution also has the potential to serve as a pre-processing module to prepare data for other sophisticated tasks (e.g., semi-supervised learning \citep{xie2019unsupervised,berthelot2019mixmatch}). %

Our main contributions are: 
\squishlist
\item \emph{New perspective:} Different from current methods that train customized models on noisy datasets, we propose a training-free and data-centric solution to efficiently detect corrupted labels\rev{, which provides a new and complementary perspective to the traditional learning-centric solution. We demonstrate the effectiveness of this simple idea and open the possibility for follow-up works.} 
\item \emph{Efficient algorithms:} Based on the neighborhood information, we propose two methods: a voting-based local detection method that only requires checking the noisy label consensuses of nearby features, and a ranking-based global detection method that scores each instance by its likelihood of being clean and filters out a guaranteed percentage of instances with low scores as corrupted ones. 
\item \emph{Theoretical analyses:} We theoretically analyze how the quality of features (but possibly imperfect in practice) affects the local voting and provide guidelines for tuning neighborhood size. We also prove the worst-case error bound for the ranking-based method. 
\item \emph{Numerical findings:} Our numerical experiments show three important messages: in corrupted label detection, i) training with noisy supervisions may not be necessary; ii) feature extraction layers tend to be more useful than the logit layers; iii) features extracted from other tasks or domains are helpful. 
\squishend

\subsection{Related Works}

\textbf{Learning with noisy labels~}
There are many other works that can detect corrupted instances (a.k.a. sample selection) in the literature, e.g., \citep{han2018co,yu2019does,yao2020searching,wei2020combating,jiang2020beyond,zhang2021learning,huang2019o2u}, and its combination with semi-supervised learning \citep{wang2020seminll,Li2020DivideMix,cheng2021learningsieve}.
Another line of works focus on designing robust loss functions to mitigate the effect of label noise, such as numerical methods \citep{ghosh2017robust,zhang2018generalized,gong2018decomposition,amid2019robust,wang2019symmetric,shu2020learning,wang2022pico} and statistical methods \citep{natarajan2013learning,liu2015classification,patrini2017making,liu2019peer,xia2019anchor,zhu2020second,jiang2022an,feng2021can,wei2021understanding,wei2022deep}.
They all require training DNNs with noisy supervisions and would suffer from the limitations of the learning-centric methods. %

\rev{\textbf{$k$-NN for noisy labels~}
The $k$-NN technique often plays important roles in building auxiliary methods to improve deep learning \cite{jiang2018trust}. Recently, it has been extended to filtering out corrupted instances when learning with noisy labels
\cite{gao2016resistance,reeve2019fast,kong2020knn,bahri2020deep}. However, these methods focus on learning-centric solutions and cannot avoid memorizing noisy labels.}

\textbf{Label aggregation~}
Our work is also relevant to the literature of crowdsourcing that focuses on label aggregation (to clean the labels) \citep{liu2012variational,karger2011iterative,karger2013efficient,liu2015online,zhang2014spectral,wei2022aggregate}. %
Most of these works can access multiple reports (labels) for the same input feature, while our real-world datasets usually have only one noisy label for each feature.

%% file: src_icml/related.tex
\section{Preliminaries}\label{sec:pre}
\textbf{Instances~}
Traditional supervised classification tasks build on a clean dataset $D:=\{(x_n,y_n)\}_{n\in[N]}$, where $[N]:= \{1,2,\cdots,N\}$. Each \emph{clean instance} $(x_n,y_n)$ includes \emph{feature} $x_n$ and \emph{clean label} $y_n$, which is drawn according to random variables $(X,Y)\sim \mathcal D$.
In many practical cases, the clean labels may be unavailable and the learner could only observe a noisy dataset denoted by $\widetilde D:=\{(x_n,\tilde y_n)\}_{n\in[N]}$, where $(x_n,\tilde y_n)$ is a \emph{noisy instance} and the \emph{noisy label} $\tilde y_n$ may or may not be identical to $y_n$.
We call $\tilde y_n$ is \emph{corrupted} if $\tilde y_n \ne y_n$ and clean otherwise.
The instance $(x_n,\tilde y_n)$ is a \emph{corrupted instance} if $\tilde y_n$ is corrupted.
The noisy data distribution corresponds to $\widetilde D$ is $(X,\widetilde Y)\sim \widetilde{\mathcal D}$.
We focus on the closed-set label noise that $Y$ and $\widetilde Y$ are assumed to be in the same label space, e.g., $Y,\widetilde Y\in[K]$.
Explorations on open-set data \citep{xia2020extended,wei2021open,luo2021empirical} are deferred to future works.

\textbf{Clusterability~}
In this paper, we focus on a setting where the distances between two features should be comparable or clusterable \citep{zhu2021clusterability}, i.e., nearby features should belong to the same true class with a high probability \citep{gao2016resistance}, which could be formally defined as:
\begin{defn}[$(k,\delta_k)$ label clusterability]\label{def:k_delta_cluster}
A dataset $D$ satisfies $(k,\delta_k)$ label clusterability if: $\forall n \in [N]$, the feature $x_n$ and its $k$-Nearest-Neighbors ($k$-NN) $x_{n_1}, \cdots, x_{n_k}$ belong to the same true class with probability at least $1-\delta_k$. 
\end{defn}

Note $\delta_k$ captures two types of randomnesses: one comes from a probabilistic $Y$ given $X$, i.e., $\exists i,x, \PP(Y=i|X=x) \notin\{0,1\}$; the other depends on the quality of features and the value of $k$, which will be further illustrated in Figure~\ref{fig:delta_prob}.
The $(k,0)$ label clusterability is also known as $k$-NN label clusterability \citep{zhu2021clusterability}.

\textbf{Corrupted label detection~}
Our paper aims to improve the performance of the corrupted label detection (a.k.a. finding label errors) which is measured by the $F_1$-score of the \textit{detected corrupted instances}, which is the harmonic mean of the precision and recall, i.e. $$F_1={2}/{({\tt Precision}^{-1} + {\tt Recall}^{-1})}.$$ Let $\BR(\cdot)$ be the indicator function that takes value $1$ when the specified condition is satisfied and $0$ otherwise. Let $v_n=1$ indicate that $\tilde y_n$ is detected as a corrupted label, and $v_n=0$ if $\tilde y_n$ is detected to be clean. Then the precision and recall can be calculated as 
\begin{align*}
    {\tt Precision} &= \frac{\sum_{n\in[N]}\BR(v_n=1, \tilde y_n \ne y_n)}{\sum_{n\in[N]}\BR(v_n=1)}, \\ {\tt Recall} & =\frac{\sum_{n\in[N]}\BR(v_n=1, \tilde y_n \ne y_n)}{\sum_{n\in[N]} \BR (\tilde y_n \ne y_n)}.
\end{align*}
\rev{Note the $F_1$ score on corrupted instances is sensitive to the case when the noise rate is mild to low, which is typically the case in practice. For example, if $20\%$ of the data is corrupted but the algorithm reports no label errors, the returned $F_1$ score on corrupted instances is $0$ while the one on clean instances is $2/(0.8^{-1} + 1) \approx 0.89$.     
}

%% file: src_icml/method.tex
\section{Corrupted Label Detection Using Similar Features}

Different from most methods that detect corrupted instances based on the logit layer or model predictions \citep{northcutt2021confident,cheng2021learningsieve,pruthi2020estimating,bahri2020deep}, we focus on a more data-centric solution that operates on features. Particularly we are interested in the possibility of detecting corrupted labels in a training-free way.
In this section, we will first introduce intuitions, and then provide two efficient algorithms to detect corrupted labels with similar features.

\subsection{Intuitions}

\begin{figure}
    \centering
    \includegraphics[width=0.35\textwidth]{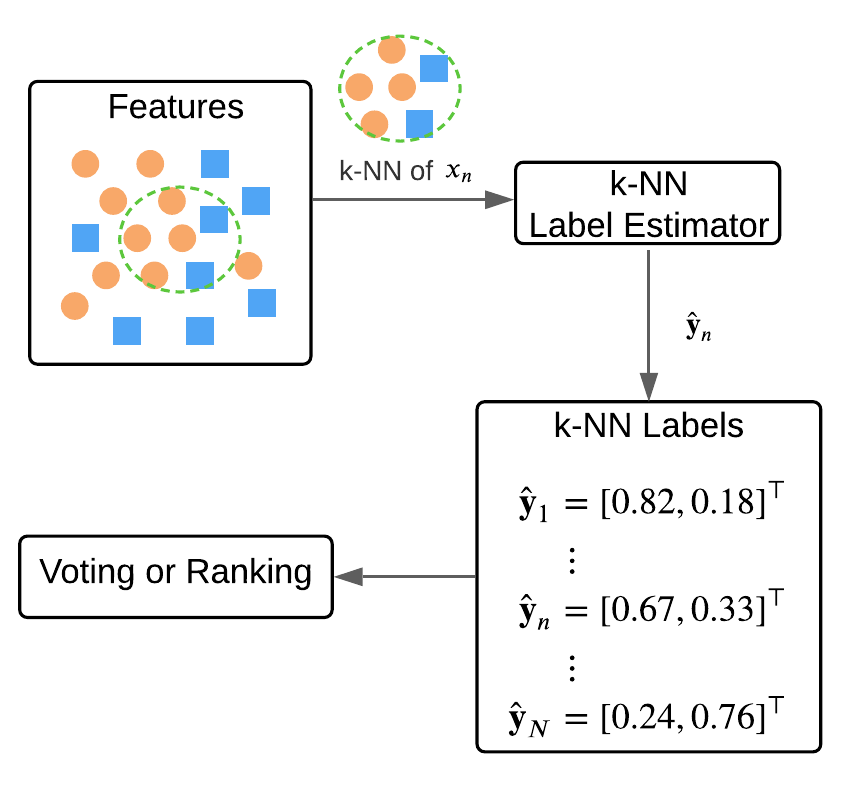}
    \caption{Detect corrupted labels with similar features.
{\color{myorange} Orange circle}: instance with noisy label $1$.
{\color{myblue} Blue square}: instance with noisy label $2$. {\color{mygreen} Green dashed circle}: A $k$-NN example.
}
    \label{fig:illustrate}
\vskip -0.2in %
\end{figure}

The learning-centric detection methods often detect corrupted instances by comparing model predictions with noisy labels \citep{cheng2021learningsieve,northcutt2021confident} as illustrated in Figure~\ref{fig:data-centric}.
However, for the data-centric method, the feature $x_n$ cannot be directly compared with the noisy label $\tilde y_n$ since $x_n$ is not directly categorical without a model, i.e., the connection between a single $x_n$ and $\tilde y_n$ is weak. Thus our first step should be establishing an auxiliary categorical information using only features.

As illustrated in Figure~\ref{fig:illustrate}, the high-level intuition is to check label consensuses of nearby features. With $(k,0)$ label clusterability as in Definition~\ref{def:k_delta_cluster}, we know the true labels of $x_n$ and its $k$-NN $x_{n_1},\cdots,x_{n_k}$ should be the same. 
\rev{If we further assume the label noise is group-dependent \citep{wang2021fair}, i.e., each $x_n$ and its $k$-NN can be viewed as a local group and share the same noise transition matrix \cite{liu2022identifiability}: 
$\PP(\widetilde Y=\tilde y_n|X=x_n, Y=y_n) = \PP(\widetilde Y=\tilde y_n|X=x_{n'}, Y=y_n), \forall x_{n'}\in \{x_{n_1},\cdots, x_{n_k}\}$,
we can first treat their noisy labels as $k+1$ independent observations of $\PP(\widetilde Y = \tilde y_n|X=x_n, Y=y_n)$,} then estimate the probability by counting the (weighted) frequency of each class in the \emph{$k$-NN label estimator}, and get \emph{$k$-NN labels} $\hat {\bm y}_n$. \rev{We use the \textbf{bold} $\bm y$ to indicate a vector, which can be seen as either an one-hot encoding of a hard label or a soft label \cite{zhu2022the}.}
The $i$-th element $\hat \vy_{n}[i]$ can be interpreted as the estimated probability of predicting class-$i$.

\rev{Note the $k$-NN technique has been implemented by \citet{bahri2020deep} as a filter to remove corrupted instances. However, this approach focuses on calculating distances on the logit layer, which inevitably requires a task-specific training process and may suffer from the limitations mentioned in Section~\ref{sec:intro}. Besides, using appropriate features may be better than model logits/predictions when the dataset is noisy. See discussions below.}

\textbf{Features could be better than model predictions~}
During supervised training, memorizing noisy labels makes the model generalizes poorly \citep{han2020survey}, while using only features may effectively avoid this issue \cite{li2021does}.
For those pre-extracted features, e.g., tabular data in UCI datasets \cite{dua2017uci}, the input features are already comparable and directly applying data-centric methods on these features avoids memorizing noisy labels.
For more challenging tasks such as image or text classifications, we can also borrow some pre-trained models to pre-process the raw feature to improve the clusterability of features, such as BERT \citep{devlin-etal-2019-bert} for language tasks, CLIP \citep{radford2021learning} for vision-language tasks, or some feature extractors from unsupervised learning \citep{ji2019invariant} and self-supervised learning \citep{jaiswal2021survey, liu2021tera, he2020momentum,chen2020simple,cheng2021demystifying}, which are not affected by noisy labels.

\subsection{Voting-Based Local Detection}

Inspired by the idea implemented in model decisions, i.e., selecting the most likely class as the true class, we can simply ``predict'' the index that corresponds to the largest element in $\hat {\bm y}_n$ with random tie-breaking, i.e., $y^{\texttt{vote}}_n = \argmax_{i\in[K]} \hat {\vy}_n[i].$ %
To further detect whether $\tilde y_n$ is corrupted or not, we only need to check $v_n:=\BR(y^{\texttt{vote}}_n \ne \tilde y_n).$ Recall $v_n=1$ indicates a corrupted label.
This voting method relies only on the local information within each $k$-NN label $\hat {\bm y}_n$, which may not be robust with low-quality features. Intuitively, when the gap between the true class probability and the wrong class probability is small, the majority vote will be likely to make mistakes due to sampling errors in $\hat \vy_n$. Thus only using local information within each $\hat {\bm y}_n$ may not be sufficient.
It is important to leverage more information such as some global statistics, which will be discussed later.

\subsection{Ranking-Based Global Detection}
\rev{The score function can be designed to detect corrupted instances \citep{northcutt2021confident,cheng2021learningsieve,pruthi2020estimating,bahri2020deep}, hard-to-learn instances \cite{liu2021just}, out-of-distribution instances \cite{wei2022mitigating}, and suspicious instances that may cause model unfairness \cite{wang2022understanding}. However, it is not clear how to do so without training a task-specific model.}
From a global perspective, if the likelihood for each instance being clean could be evaluated by some scoring functions, we can sort the scores in an increasing order and filter out the low-score instances as corrupted ones. Based on this intuition, there are two critical components: the \emph{scoring function} and the \emph{threshold} to differentiate the low-score part (corrupted) and the high-score part (clean). 

\textbf{Scoring function~}
A good scoring function should be able to give clean instances higher scores than corrupted instances. 
We adopt cosine similarity defined as: 
$$\textsf{Score}(\hat \vy_n,j) =  \frac{\hat \vy_n^\top \ve_j }{\|\hat \vy_n\|_2 \|\ve_j \|_2},$$
where $\bm e_j$ is the one-hot encoding of label $j$.
To evaluate whether the soft label $\hat \vy_n$ informs us a clean instance or not, we compare $\textsf{Score}(\hat \vy_n,\tilde y_n)$ with other instances that have the same noisy label.
This scoring function captures more information than majority votes, which is summarized as follows.

\begin{property}[Relative score]\label{property:rel}
Within the same instance, the score of the majority class is higher than the others, \rev{i.e., $\forall j\ne y^{\texttt{vote}}_n, j \in [K], \forall n\in[N]:$ $\textsf{Score}(\hat \vy_n,y^{\texttt{vote}}_n) > \textsf{Score}(\hat \vy_n,j).$}
\end{property}

\begin{property}[Absolute score]\label{property:abs}
$\textsf{Score}(\hat \vy_n,j)$ is jointly determined by both $\hat \vy_n[j]$ and $\hat \vy_n[j'], \forall j'\ne j$.
\end{property}

The first property guarantees that the corrupted labels would have lower scores than clean labels for the same instance when the vote is correct.
However, although solely relying on Property~\ref{property:rel} may work well in the voting-based method which makes decisions individually for each instance, it is not sufficient to be trustworthy in the ranking-based global detection.
\rev{Empirically we observe that if we choose a score function that Property 3.2 does not hold, e.g., treating $k$-NN soft labels as model predictions and check the cross-entropy loss, it does not always return satisfying results in our experiments.}
The main reason is that, across different instances, the non-majority classes of some instances may have higher absolute scores than the majority classes of the other instances, which is especially true for general instance-dependent label noise with heterogeneous noise rates \citep{cheng2021learningsieve}.
Property~\ref{property:abs} helps make it less likely to happen.
Consider an example as follows.

\textbf{Example~}
Suppose $\hat \vy_{n_1}=\hat \vy_{n_2}=[0.6,0.4,0.0]^\top$, $\hat \vy_{n_3}=[0.34,0.33,0.33]^\top$,  $y_{n_1}=y_{n_2}=y_{n_3}=1$, $\tilde y_{n_1}=\tilde y_{n_3}=1, \tilde y_{n_2} = 2$.
We can use the majority vote to get perfect detection in this case, i.e., $y^{\texttt{vote}}_{n_1}=y^{\texttt{vote}}_{n_2}=y^{\texttt{vote}}_{n_3}=1=y_{n_1}$, since the first class of each instance has the largest value.
However, if we directly use a single value in soft label $\vy_{n}$ to score them, e.g., $\textsf{Score}'(\hat \vy_n,j)=\hat \vy_{n}[j]$, we will have $\hat \vy_{n_1}[\tilde y_{n_1}] =0.6 > \hat \vy_{n_2}[\tilde y_{n_2}] = 0.4 > \hat \vy_{n_3}[\tilde y_{n_3}]=0.34$, where the ranking is $n_3\prec n_2\prec n_1$.
Ideally, we know instance $n_2$ is corrupted and the true ranking should be $n_2 \prec n_3 \prec n_1$ or $n_2  \prec n_1 \prec n_3$.
To mitigate this problem, we choose the cosine similarity as our scoring function. The three instances could be scored as $0.83,0.55,0.59$, corresponding to an ideal ranking $n_2 \prec n_3 \prec n_1$. We formally introduce the detailed ranking approach as follows.

\textbf{Ranking~} Suppose we have a group of instances with the same noisy class $j$, i.e. $\{(x_n,\tilde y_n)\}_{n\in\mathcal N_j}$, where $\mathcal N_j :=\{n|\tilde y_n = j\}$ are the set of indices that correspond to noisy class $j$. Let $N_j$ be the number of indices in $\mathcal N_j$ (counted from noisy labels).
Intuitively, we can first sort all instances in $\mathcal N_j$ in an increasing order by \texttt{argsort} and obtain the original indices for the sorted scores as:
$$\mathcal I = \texttt{argsort}\{\textsf{Score}(\hat \vy_n,j)\}_{n\in \mathcal N_j},$$ where the low-score head is supposed to consist of corrupted instances \citep{northcutt2021confident}.
Then we can simply select the first $\widetilde N_j$ instances with low scores as corrupted instances:
$$
v_n = \BR(\texttt{Loc}(n,\mathcal I) \le \widetilde{N}_j),
$$
where $\texttt{Loc}(n,\mathcal I)$ returns the index of $n$ in $\mathcal I$.
Instead of manually tuning $\widetilde N_j$ \cite{han2018co}, we discuss how to determine it algorithmically.

\textbf{Threshold~}
The number of corrupted instances in $\mathcal N_j$ is approximately $\PP(Y\ne j|\widetilde Y=j) \cdot N_j$ when $N_j$ is sufficiently large. Therefore if all the corrupted instances have lower scores than any clean instance, we can set $\widetilde N_j = \PP(Y\ne j|\widetilde Y=j) \cdot N_j$ to obtain the ideal division.
Note $N_j$ can be obtained by directly counting the number of instances with noisy label $j$. 
To calculate the probability $$\PP(Y\ne j|\widetilde Y=j) = 1- \PP(Y = j|\widetilde Y=j),$$ we borrow the results from the HOC estimator \citep{zhu2021clusterability,zhu2022beyond}, where the noise transition probability $\PP(\widetilde Y = j|Y = j)$ and the marginal distribution of clean label $\PP(Y=j)$ can be estimated with only features and the corresponding noisy labels.
Then we can calculated our needed probability by Bayes' rule $$\PP(Y=j|\widetilde Y=j) = \PP(\widetilde Y=j|Y=j) \cdot \PP(Y=j)/\PP(\widetilde Y=j),$$ where $\PP(\widetilde Y=j)$ can be estimated by counting the frequency of noisy label $j$ in $\widetilde{D}$.
Technically other methods exist in the literature to estimate $\PP(\widetilde Y|Y)$ \citep{liu2015classification,patrini2017making,northcutt2021confident,li2021provably}. But they often require training a model to fit the data distribution, which conflict with our goal of a training-free solution; instead, HOC fits us perfectly.

{%
\begin{algorithm*}[t]

   \caption{Detection with {\bf Simi}lar {\bf Feat}ures (The \ours{} Detector)}
   \label{alg:all}
{%
\begin{algorithmic}[1]
   \STATE {\bfseries Input:} Number of epochs: $M$. $k$-NN parameter: $k$. Noisy dataset: $\widetilde{D}=\{( x_n, \tilde y_n)\}_{n\in  [N]}$. Feature extractor: $g(\cdot)$. Method: \textit{Vote} or \textit{Rank}. Epoch counter $m=0$.
    \REPEAT \label{line:rndsmp_0}
    \STATE $x'_n\leftarrow$ \texttt{RandPreProcess}$(x_n), \forall n$;  \hfill
    \algcom{\# Initialize \& Standard data augmentations }
    \\
    \STATE $x_n \leftarrow g(x'_n), \forall n$; \hfill
    \algcom{\# For tasks with rarely clusterable features, extract features with $g(\cdot)$}
    \\
    \STATE $\hat {\bm y}_n \leftarrow$  \texttt{kNNLabel}($\{x_n\}_{n\in[N]}, k$) \hfill \algcom{\# Get soft labels. One can weight instances by the similarity to the center instance.
    }
    \IF{\textit{Vote}} 
    \STATE 
    $y^{\texttt{vote}}_n \leftarrow \argmax_{i\in[K]} \hat {y}_n[i]$;
    \hfill \algcom{\# Apply local majority vote} \label{alg:vote1}
    \STATE
    $v_n\leftarrow\BR(y^{\texttt{vote}}_n \ne \tilde y_n), \forall n\in[N]$; \hfill \algcom{\# Treat as corrupted if majority votes disagree with noisy labels}\label{alg:vote2}
    \ELSE
    \STATE    $\PP(Y), \PP(\widetilde Y|Y)\leftarrow\texttt{HOC}(\{(x_n,\tilde y_n)\}_{n\in[N]})$; %
    \hfill \algcom{\# Estimate clean priors $\PP(Y)$ and noise transitions $\PP(\widetilde Y|Y)$ by HOC}
    \STATE $\PP(Y |\widetilde Y) = \PP(\widetilde Y|Y) \cdot \PP(Y)/\PP(\widetilde Y)$; 
    \hfill \algcom{\# Estimate thresholds by Bayes' rule}
    \FOR{$j$ {\bf in} $[K]$}
    \STATE $\mathcal N_j :=\{n|\tilde y_n = j\}$; \hfill \algcom{\# Detect corrupted labels in each set $\mathcal N_j$}
    \STATE $\mathcal I \leftarrow\texttt{argsort}\{\textsf{Score}(\hat \vy_n,j)\}_{n\in \mathcal N_j}$; \hfill \algcom{\# $\mathcal I$ records the raw index of each sorted value}\label{alg:rank1}
    \STATE $v_n \leftarrow \BR\big(\texttt{Loc}(n,\mathcal I) \le \lfloor (1-\PP(Y=j|\widetilde Y=j)) \cdot N_j \rfloor\big)$; 
    \hfill \algcom{\# Select low-score (head) instances as corrupted ones}\label{alg:rank2}
    \ENDFOR
    \ENDIF
    \STATE $\mathcal V_m = \{v_n\}_{n\in[N]}$; \hfill \algcom{\# Record detection results in the $m$-th epoch}
    \UNTIL{$M$ times} \label{line:rndsmp_1}
    \STATE $\mathcal V = $\texttt{Vote}($\mathcal V_m, \forall m \in [M]$); \hfill \algcom{\# Do majority vote based on results from $M$ epochs} \label{line:vote-epoch}
    \STATE {\bfseries Output:} $[N]\setminus \mathcal V$.
\end{algorithmic}}
\end{algorithm*}}

\subsection{Algorithm: \ours{}}
Algorithm~\ref{alg:all} summarizes our solution.
The main computation complexity is pro-processing features with extractor $g(\cdot)$, which is less than the cost of evaluating the model compared with the training-based methods.
Thus \ours{} can filter out corrupted instances efficiently.
In Algorithm~\ref{alg:all}, we run either voting-based local detection as Lines~\ref{alg:vote1},~\ref{alg:vote2},
or ranking-based global detection as Lines~\ref{alg:rank1},~\ref{alg:rank2}. 
The detection is run multiple times with random standard data augmentations to reduce the variance of estimation. 
The majority of results from different epochs is adopted as the final detection output as Line~\ref{line:vote-epoch}, i.e., flag as corrupted if $v_n=1$ in more than half of the epochs.

%% file: src_icml/analyses.tex
\section{How Does Feature Quality Affect Our Solution?}

In this section, we will first show how the quality of features\footnote{Note the voting-based method achieves an $F_1$-score of $1$ when $(k,0)$-NN label clusterability, $k\rightarrow +\infty$, holds.} affects the selection of the hyperparameter $k$, then analyze the error upper bound for the ranking-based method.

\subsection{How Does Feature Quality Affect the Choice of $k$?}

Recall $k$ is used as illustrated in Figure~\ref{fig:illustrate}.
On one hand, the $k$-NN label estimator will be more accurate if there is stronger clusterability that more neighbor features belong to the same true class \citep{liu2015online,zhu2021clusterability}, which helps improve the performance of later algorithms.
On the other hand, with good but imperfect features, stronger clusterability with a larger $k$ is less likely to satisfy, thus the violation probability $\delta_k$ increases with $k$ for a given extractor $g(\cdot)$.
We take the voting-based method as an example and analyze this tradeoff.
For a clean presentation, we focus on a binary classification with instance-dependent label noise where $\PP(Y=1) = p$, $\PP(\widetilde Y=2 |X, Y=1) = e_1(X)$, $\PP(\widetilde Y=1|X,Y=2) =  e_2(X)$. 
Suppose the instance-dependent noise rate is upper-bounded by $e$, i.e., $e_1(X) \le e, e_2(X)\le e$.
With $\delta_k$ as in Definition~\ref{def:k_delta_cluster}, we calculate the lower bound of the probability that the vote is correct in Proposition~\ref{thm:vote}.

\begin{prop}\label{thm:vote}
The lower bound for the probability of getting true detection with majority vote is $$\PP(\text{Vote is correct}|k) \ge  {(1-\delta_{k})} \cdot {I_{1-e}(k+1-k',k'+1)},$$
where $k'=\lceil (k+1)/2 \rceil -1$, $I_{1-e}(k+1-k',k'+1)$ is the regularized incomplete beta function defined as
$
I_{1-e}(k+1-k',k'+1) = (k+1-k'){k+1\choose k'} \int_0^{1-e} t^{k-k'}(1-t)^{k'} dt.
$
\end{prop}
\begin{figure*}
    \centering
    \includegraphics[width=.7\textwidth]{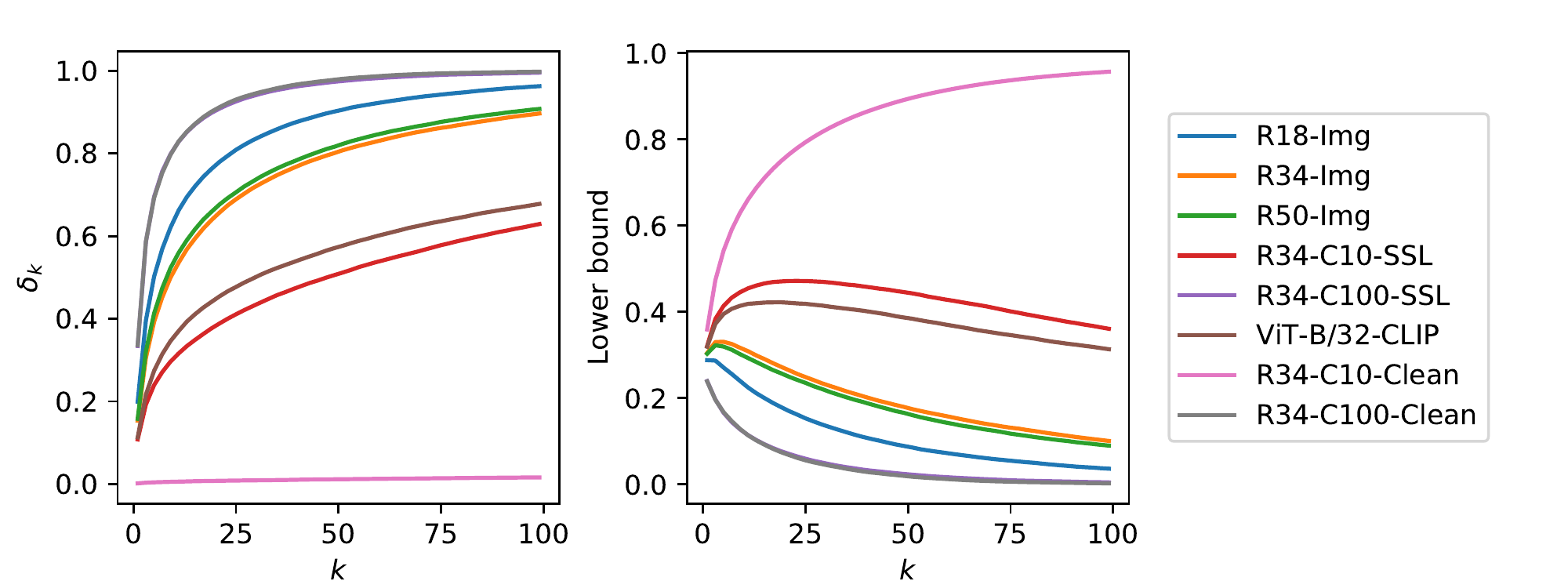}
    \caption{The trends of $\delta_k$ and probability lower bounds on CIFAR-10 \citep{krizhevsky2009learning} when raw features are extracted with different $g(\cdot)$. The outputs of the last convolution layer are adopted. R18/34/50: ResNet18/34/50. Img: Pre-trained on ImageNet \citep{imagenet_cvpr09}. C10/100-Clean: Pre-trained on clean CIFAR-10/100. C10/100-SSL: Pre-trained on CIFAR-10/100 without labels by SimCLR \citep{chen2020simple}. ViT-B/32-CLIP: CLIP Pre-trained vision transformer \citep{radford2021learning}.}
    \label{fig:delta_prob}
\vskip -0.1in %
\end{figure*}
Proposition~\ref{thm:vote} shows the tradeoff between a reliable $k$-NN label and an accurate vote.
When $k$ is increasing, {\bf Term-1} $(1-\delta_k)$ (quality of features) decreases but {\bf Term-$2$} ${I_{1-e_1}(k+1-k',k'+1)}$ (result of pure majority vote) increases.
With Proposition~\ref{thm:vote}, we are ready to answer the question: \emph{when do we need more labels?}  See Remark~\ref{rmk:k}. %
\begin{rmk}\label{rmk:k}
Consider the lower bounds with $k_1$ and $k_2$ ($k_1 < k_2$).
Supposing the first lower bound is lower than the second lower bound, based on Proposition~\ref{thm:vote}, we roughly study the trend with an increasing $k$ by comparing two bounds and get
\begin{equation*}
    \begin{split}
        \frac{1-\delta_{k_1}}{1-\delta_{k_2}} <  \frac{I_{1-e}(k_2+1-k'_2,k'_2+1)}{I_{1-e}(k_1+1-k'_1,k'_1+1)}.
    \end{split}
\end{equation*}
For example, when $k_1=5$, $k_2=20$, $e=0.4$, we can calculate the incomplete beta function and $\frac{1-\delta_{5}}{1-\delta_{20}} < 1.52$. Supposing $\delta_{5}=0.2$, we have $\delta_{20}<0.47$. This indicates increasing $k$ from $5$ to $20$ would not improve the lower bound with features satisfying $\delta_{20}>0.47$.
This observation helps us set $k$ with practical and imperfect features. We set $k=10$ in all of our experiments. 
\end{rmk}

Remark~\ref{rmk:k} indicates that: {\bf \emph{with practical (imperfect) features, a small $k$ may achieve the best (highest) probability lower bound}}. To further consolidate this claim, we numerically calculate $\delta_k$ with different quality of features on CIFAR-10 and the corresponding probability lower bound in Figure~\ref{fig:delta_prob}. We find most of the probability lower bounds first increase then decrease except for the ``perfect'' feature which is extracted by the extractor trained using ground-truth labels.
Note this feature extractor has memorized all clean instances so that $\delta_k \rightarrow 0$ since $k \ll 5000$ (the number of instances in the same label class).

\subsection{How Does Feature Quality Affect $F_1$-Score?}\label{sec:thm1}

We next prove the probability bound for the performance of the ranking-based method.
Consider a $K$-class classification problem with informative instance-dependent label noise \citep{cheng2021learningsieve}.
Denote random variable $S$ by the score of each instance being clean. A higher score $S$ indicates the instance is more likely to be clean.
\rev{
Denote the score of a true/false instance by $S^{\text{true}}_{n,j}:=\textsf{Score}(\hat \vy_n,j)$ when $\tilde y_n = y_n = j$ and $S^{\text{false}}_{n',j}:=\textsf{Score}(\hat \vy_{n'},j)$ when $\tilde y_{n'} = j, y_{n'} \ne j$.
Both are scalars.
Then for instances in $\mathcal N_j$, we have two set of random variables $\mathbb S_j^{\text{true}}:=\{S^{\text{true}}_{n,j}|n\in\mathcal N_j, \tilde y_n = y_n = j\}$ and $\mathbb S_j^{\text{false}}:=\{S^{\text{false}}_{n',j}|n'\in\mathcal N_j,\tilde y_{n'} =j, y_{n'} \ne j\}$.
Recall $\mathcal N_j :=\{n|\tilde y_n = j\}$ are the set of indices that correspond to noisy class $j$.
Intuitively, the score $S_{n,j}^{\text{true}}$ should be greater than $S_{n',j}^{\text{false}}$.
Suppose their means, which depend on noise rates, are bounded, i.e., $$\mathbb E[S_{n,j}^{\text{true}}] \ge \mu_j^{\text{true}}, \quad \mathbb E[S_{n',j}^{\text{false}}] \le \mu_j^{\text{false}}$$ for all feasible $n, n'$.
Assume there exists a feasible $v$ such that
both $S_j^{\text{true}}$  and $S_j^{\text{false}}$ follow sub-Gaussian distributions with variance proxy $\frac{\Delta^2}{2v}$ \citep{buldygin1980sub,zhu2021federated} such that:
$$\PP( \mu_j^{\text{true}} - S_{n,j}^{\text{true}} \ge t) \le e^{-\frac{vt^2}{\Delta^2}}, \PP(S_{n',j}^{\text{false}} - \mu_j^{\text{false}} \ge  t) \le e^{-\frac{vt^2}{\Delta^2}},$$ and the probability density satisfies $\PP(S_j^{\text{true}} = \mu_j^{\text{true}} ) = \PP(S_j^{\text{false}} = \mu_j^{\text{false}} ) = 1/\Delta$, where $1/\Delta$ is the ``height'' of both distributions, $v$ is the decay rate of tails.
Let $N_j^-$ ($N_j^+$) be the number of indices in $\mathbb S_j^{\text{false}}$ ($\mathbb S_j^{\text{true}}$).
Theorem~\ref{thm:rank} summarizes the performance bound of the ranking-based method. See Appendix for the proof.}

\begin{thm}\label{thm:rank}
With probability at least $p$, \rev{when the threshold for the ranking-based method is set to $$(1-\PP(Y=j|\widetilde Y=j)) \cdot N_j$$ as Line~15, the $F_1$-score of detecting corrupted instances in $\mathcal N_j$ by ranking is at least $1-\frac{e^{-{v}} \max( N^-, N^+ ) + \alpha}{N^-}$,} where $p=\int_{-1}^{\mu^{\text{true}}_j - \mu^{\text{false}}_j -\Delta} f(t) dt$, $f(t)$ is the probability density function of the difference of two independent beta-distributed random variables $\beta_1-\beta_2$, where $\beta_1\sim Beta(N^-,1), \beta_2\sim Beta(\alpha+1,N^+-\alpha)$.
\end{thm}
Theorem~\ref{thm:vote} shows the detection performance depends on:
\vspace{-.5cm}
\rev{\squishlist
\item the concentration of $S^{\text{true}}_{n,j}$ and $S^{\text{false}}_{n',j}$: variance proxy $\frac{\Delta^2}{2v}$;
\item the distance between $S^{\text{true}}_{n,j}$ and $S^{\text{false}}_{n',j}$: $\mu^{\text{true}}_j - \mu^{\text{false}}_j$.
\squishend}
Intuitively, with proper scoring function and high-quality features, we have small variance proxy (small $\Delta$ and large $v$) and $F_1$-score approximates to $1$.

%% file: src_icml/exp.tex
\section{Empirical Results}\label{sec:exp}

We present experimental evidence in this section.
The performance is measured by the $F_1$-score of the detected corrupted labels as defined in Section~\ref{sec:pre}.
Note there is no training procedure in our method.
The only hyperparameters in our methods are the number of epochs $M$ and the $k$-NN parameter $k$.
Intuitively, a larger $M$ returns a collective result from more times of detection, which should be more accurate.
\rev{But a larger $M$ takes more time. We set $M=21$ (an odd number for better tie-breaking) for an efficient solution.}
The hyperparameter $k$ cannot be set too large as demonstrated in Figure~\ref{fig:delta_prob}.
\rev{From Figure 3, we notice that the lower bound (RHS figure) is relatively high when $k=10$ for all settings.
Therefore, in CIFAR \citep{krizhevsky2009learning} experiments, rather than fine-tune $M$ and $k$ for different settings, we fix $M=21$ and $k=10$.}
We also test on Clothing1M \citep{xiao2015learning}. Detailed experiment settings on Clothing1M are in Appendix~\ref{supp:c1m}.

\textbf{Synthetic label noise~}
We experiment with three popular synthetic label noise models: the \emph{symmetric} label noise, the \emph{asymmetric} label noise, and the \emph{instance-dependent} label noise.
Denote the ratio of instances with corrupted labels in the whole dataset by $\eta$.
Both the symmetric and the asymmetric noise models follow the class-dependent assumption \citep{liu2015classification}, i.e., the label noise only depends only on the clean class: $\PP(\widetilde Y|X,Y) = \PP(\widetilde Y|Y)$.
Specially, the symmetric noise is generated by uniform flipping, i.e., randomly flipping a true label to the other possible classes w.p. $\eta$ \citep{cheng2021learningsieve}.
The asymmetric noise is generated by pair-wise flipping, i.e., randomly flipping true label $i$ to the next class $(i \mod K) + 1$. 
Denote by $d$ the dimension of features.
The instance-dependent label noise is synthesized by randomly generating a $d\times K$ projection matrix $w_i$ for each class $i$ and project each incoming feature with true class $y_n$ onto each column of $w_{y_n}$ \citep{xia2020parts}. Instance $n$ is more likely to be flipped to class $j$ if the projection value of $x_n$ on the $j$-th column of $w_{y_n}$ is high. See Appendix~B in \cite{xia2020parts} and Appendix~D.1 in \cite{zhu2021clusterability} for more details. %
We use symmetric noise with $\eta=0.6$ (\emph{Symm. 0.6}), asymmetric noise with $\eta=0.3$ (\emph{Asym. 0.3}), and instance-dependent noise with $\eta=0.4$ (\emph{Inst. 0.4}) in experiments.

\textbf{Real-world label noise~}
The \emph{real-world} label noise comes from human annotations or weakly labeled web data.
We use the $50,000$ noisy training labels ($\eta \approx 0.16$) for CIFAR-10 collected by \cite{zhu2021clusterability}, and $50,000$ noisy training labels ($\eta \approx 0.40$) for CIFAR-100 collected by \cite{wei2022learning}.
Both sets of noisy labels are crowd-sourced from Amazon Mechanical Turk.
{For Clothing1M \citep{xiao2015learning}, we could not calculate the $F_1$-scores due to the lack of ground-truth labels. We firstly perform noise detection on 1 million noisy training instances then train only with the selected clean data to check the effectiveness.}

\subsection{Fitting Noisy Distributions May Not Be Necessary}\label{sec:supervision_not}

\begin{table*}[!t]
	\caption{Comparisons of $F_1$-scores (\%). CORES, CL, TracIn: Train with noisy supervisions. \ours{}-V and \ours{}-R: Get $g(\cdot)$ \textbf{without} any supervision. Top 2 are {\bf bold}.}
	\label{table:supervision}
\begin{center}
    \begin{small}
    \begin{sc}
	\begin{center}
		\scalebox{.9}
		{\footnotesize{\begin{tabular}{c|cccccccccc} 
			\hline 
			\multirow{2}{*}{Method}  & \multicolumn{4}{c}{{CIFAR10}} & \multicolumn{4}{c}{{CIFAR100}} &  \\
			   &\emph{Human} & \emph{Symm. 0.6}& \emph{Asym. 0.3} & \emph{Inst. 0.4} & \emph{Human} & \emph{Symm. 0.6}& \emph{Asym. 0.3}& \emph{Inst. 0.4} \\
			\hline\hline
		   	CORES &65.00 &92.94 &7.68 &\textbf{87.43}   &3.52 &\textbf{92.34}& 0.02&9.67 \\
		   	CL    &55.85 &80.59 &76.45 &62.89    &64.58 &78.98& 52.96&50.08\\
		    TracIn&55.02 &76.94 &73.47 &58.85    &61.75 &76.74& 48.42&49.89 \\
		    Deep $k$-NN & 56.21 & 82.35 & 75.24 & 63.08 & 57.40 & 70.69 & 56.75 & 63.85 \\
		   	\ours{}-V& \textbf{82.30} & \textbf{93.21} & \textbf{82.52} & 81.09 & \textbf{73.19} & 84.48 &\textbf{65.42}&\textbf{74.26} \\
		   	\ours{}-R& \textbf{83.28} & \textbf{95.56} & \textbf{83.58} & \textbf{82.26} & \textbf{74.67}& \textbf{88.68} &\textbf{62.89}&\textbf{73.53} \\
		   \hline\hline
		\end{tabular}}}
	\end{center}
\end{sc}
    \end{small}
\end{center}
\vskip -0.1in %
\end{table*}

\begin{table*}[!t]
\caption{Comparisons of $F_1$-scores (\%). CORES, CL, TracIn: Use logit layers. \ours{}-V/R: Use only representations. \textbf{All methods use the same \rev{fixed} extractor from CLIP.} Top 2 are {\bf bold}.}
\label{table:rep_vs_logit}
\begin{center}
    \begin{small}
    \begin{sc}
			\scalebox{.9}
		{\footnotesize{\begin{tabular}{c|cccccccccc} 
				\hline 
				\multirow{2}{*}{Method}  & \multicolumn{4}{c}{{CIFAR10}} & \multicolumn{4}{c}{{CIFAR100}} &  \\
				   &\emph{Human} & \emph{Symm. 0.6}& \emph{Asym. 0.3} & \emph{Inst. 0.4} & \emph{Human}& \emph{Symm. 0.6}& \emph{Asym. 0.3}& \emph{Inst. 0.4} \\
				\hline\hline
				CE Sieve &67.21 &94.56 &5.24 &8.41   &16.24 & 88.55& 2.6&1.63 \\
			   	CORES    &83.18 &\textbf{96.94} &12.05 &\textbf{88.89}   &38.52 &\textbf{92.33}& 7.02&\textbf{85.52}\\
			   	CL      &69.76 &95.03 &77.14 &62.91   &67.64 &85.67& 62.58&61.53 \\
			   	TracIn  &81.85 &95.96 &80.75 &64.97   &\textbf{79.32} &\textbf{91.03}& 63.12&64.31 \\
			   	Deep $k$-NN & 82.98 & 87.47 & 76.96 & 77.42 & 72.33 & 82.95 & 64.96 & 74.25 \\
			   	\ours{}-V& \textbf{87.43} &96.44 & \textbf{88.97}&87.11 &76.26 & 86.88 & \textbf{73.50}& \textbf{80.03}\\
			   	\ours{}-R& \textbf{87.45} &\textbf{96.74} & \textbf{89.04} &\textbf{91.14} &\textbf{79.21} & 90.54 & \textbf{68.14}&77.37 \\
			   \hline\hline
			\end{tabular}}}
\end{sc}
    \end{small}
\end{center}
\vskip -0.15in %
\end{table*}

\begin{table*}[t]
\caption{Comparisons of $F_1$-scores  (\%) using $g(\cdot)$ with different $\delta_k$ (\%). Model names are the same as Figure~\ref{fig:delta_prob}. }
\label{table:diff_pretrain}
\vskip 0.1in %
	\begin{center}
	\begin{small}
    \begin{sc}
		\scalebox{.90}
		{\footnotesize{\begin{tabular}{c|cccccccc} 
			\hline 
			\multirow{2}{*}{Pre-trained Model}  & \multicolumn{3}{c}{{CIFAR10}} & 
			\multicolumn{3}{c}{{CIFAR100}} \\ 
			   & $1-\delta_k$ &\emph{Human} & \emph{Inst. 0.4} 
			   & $1-\delta_k$ &\emph{Human}& \emph{Inst. 0.4}\\
			\hline\hline
			R18-Img &35.73& 75.40 &80.22 & 11.30& 74.91&71.99 \\
		   	R34-Img &48.13& 79.52& 82.43& 16.17 &76.88&74.00 \\
		   	R50-Img & 45.77&  78.40 & 82.06  &15.81&76.55 &73.51 \\
		   	ViT-B/32-CLIP &64.12& 87.45 & 91.14 &19.94&  79.21&77.37 \\
		   	\hline
		   	R34-C10-SSL & 69.31& 83.28 & 85.26 & 2.59& 68.03& 65.94 \\
		   	R34-C10-Clean &99.41& 98.39 & 98.59 &0.22&  60.90&60.73 \\
		   	\hline
		   	R34-C100-SSL &18.59& 59.96 & 74.99  &22.46& 74.67&73.53 \\
            R34-C100-Clean &18.58&60.17 & 76.41 &89.07& 92.87& 95.29\\
		   \hline\hline
		\end{tabular}}}
\end{sc}
    \end{small}
	\end{center}
\vskip -0.1in %
\end{table*}

\begin{table}[t]
\caption{{Experiments on Clothing1M. None: Standard training with 1M noisy data. R50-Img (or ViT-B/32-CLIP, R50-Img Warmup-1): Apply our method with ResNet50 pre-trained on ImageNet (or ViT-B/32 pre-trained by CLIP, R50-Img with 1-epoch warmup). The clean test accuracy on the best epoch, the last 10 epochs, and the last epoch, are listed. Top-1 is {\bf bold}.}}
\vskip 0.1in %
\label{table:c1m}
\begin{center}
    \begin{small}
    \begin{sc}
\scalebox{.7}{\footnotesize
{
\begin{tabular}{c|cccc}
\hline
Data Selection             & \# Training & Best Epoch & Last 10                     & Last \\ \hline\hline
None                       & 1M (100\%)          & 70.32      & 69.44 $\pm$ 0.13 & 69.53      \\ 
R50-Img          & 770k (77.0\%)       & 72.37      & 71.95 $\pm$ 0.08 & 71.89      \\ 
ViT-B/32-CLIP              & 700k (70.0\%)       & 72.54      & 72.23 $\pm$ 0.17 & 72.11      \\ 
R50-Img Warmup-1 & 767k (76.7\%)       & {\bf 73.64}      & {\bf 73.28 $\pm$ 0.18} & {\bf 73.41}      \\ \hline\hline
\end{tabular}}}
\end{sc}
    \end{small}
	\end{center}
\vskip -0.15in %
\end{table}

Our first experiment aims to show that fitting the noisy data distribution may not be necessary in detecting corrupted labels.
To this end, we compare our methods, i.e., voting-based local detection (\ours{}-V) and ranking-based global detection (\ours{}-R), with three learning-centric noise detection works: CORES \citep{cheng2021learningsieve}, confident learning (CL) \citep{northcutt2021confident}, TracIn \citep{pruthi2020estimating}\rev{, and deep $k$-NN \cite{bahri2020deep}.} We use ResNet34 \cite{he2016deep} as the backbone network in this experiment.

\textbf{Baseline settings~}
All these three baselines require training a model with the noisy supervision.
Specifically, CORES \citep{cheng2021learningsieve} trains ResNet34 on the noisy dataset and uses its proposed sample sieve to filter out the corrupted instances. We adopt its default setting during training and calculate the $F_1$-score of the sieved out corrupted instances.
Confident learning (CL) \citep{northcutt2021confident} detects corrupted labels by firstly estimating probabilistic thresholds to
characterize label noise, ranking instances based on model predictions, then filtering out corrupted instances based on ranking and thresholds.
We adopt its default hyper-parameter setting to train ResNet34.
TracIn \citep{pruthi2020estimating} detects corrupted labels by evaluating the self-influence of each instance, where the corrupted instances tend to have a high influence score. The influence scores are calculated based on gradients of the last layer of ResNet34 at epoch $40, 50, 60, 100$, where the model is trained with a batch size of $128$. The initial learning rate is $0.1$ and decays to $0.01$ at epoch $50$.
Note TracIn only provides ranking for instances. To exactly detect corrupted instances, thresholds are required.
For a fair comparison, we refer to the thresholds learned by confident learning \citep{northcutt2021confident}.
Thus the corrupted instances selected by TracIn are based on the ranking from its self-influence and thresholds from CL.
To highlight that our solutions work well without any supervision, our feature extractor $g(\cdot)$ comes from the ResNet34 pre-trained by SimCLR \citep{chen2020simple} where contrastive learning is applied and \emph{no supervision} is required.
Extractor $g(\cdot)$ is obtained with only in-distribution features, e.g., for experiments with CIFAR-10, $g(\cdot)$ is pre-trained with features only from CIFAR-10.
\rev{The detailed implementation for deep $k$-NN filter \cite{bahri2020deep} is not public. Noting their $k$-NN approach is employed on the logit layer, we reproduce their work by firstly training the model on the noisy data then substituting the model logits for $g(\cdot)$ in \ours{}-V.  The best epoch result for deep $k$-NN is reported.}

\textbf{Performance~}
Table~\ref{table:supervision} compares the results obtained with or without supervisions. We can see both the voting-based and the ranking-based method achieve overall higher $F_1$-scores compared with the other three results that require learning with noisy supervisions. Moreover, in detecting the real-world human-level noisy labels, our solution outperforms baselines around $20\%$ on CIFAR-10 and $10\%$ on CIFAR-100, which indicates the training-free solution are more robust to complicated noise patterns. One might also note that CORES achieves exceptionally low $F_1$-scores on CIFAR-10/100 with asymmetric noise and CIFAR-100 with human noise. This observation also informs us that customized training processes might not be universally applicable.

\subsection{Features May Be Better Than Model Predictions}

Our next experiment aims to compare the performance of the data-centric method with the learning-centric method when the same feature extractor is adopted. Thus in this experiment, all methods adopt the same fixed feature extractor (ViT-B/32 pre-trained by CLIP \citep{radford2021learning}). Our proposed data-centric method directly operates on the extracted features, while the learning-centric method further train a linear layer with noisy supervisions based on the extracted features.
In addition to the baselines compared in Section~\ref{sec:supervision_not}, we also compare to CE Sieve \citep{cheng2021learningsieve} which follows the same sieving process as CORES but uses CE loss without regularizer. 
Other settings are the same as those in Section~\ref{sec:supervision_not}.

Table~\ref{table:rep_vs_logit} summarizes the results of this experiment. 
By counting the frequency of reaching top-2 $F_1$-scores, we find \ours{}-R wins 1st place, \ours{}-V and CORES are tied for 2nd place. However, similar to Table~\ref{table:rep_vs_logit}, we find the training process of CORES to be unstable. For instance, it almost fails for CIFAR-100 with asymmetric noise.
\rev{Besides, comparing deep $k$-NN with \ours-V, we find using the model logits given by an additional linear layer fine-tuned with noisy supervisions cannot always help improve the performance of detecting corrupted labels.}
It is therefore reasonable to believe both methods that directly deal with the extracted features achieve an overall higher $F_1$-score than other learning-centric methods.

\subsection{The Effect of the Quality of Features}
Previous experiments demonstrate our methods overall outperform baselines with high-quality features. It is interesting to see how lower-quality features perform. 
We summarize results of \ours{}-R in Table~\ref{table:diff_pretrain}.
There are several interesting findings:
1) The ImageNet pre-trained models perform well, indicating the traditional supervised training on out-of-distribution data helps obtained high-quality features;
2) For CIFAR-100, extractor $g(\cdot)$ obtained with only features from CIFAR-10 (R34-C10-SSL) performs better than the extractor with clean CIFAR-10 (R34-C10-Clean), indicating that contrastive pre-training has better generalization ability to out-of-distribution data than supervised learning;
3) The $F_1$-scores achieved by $g(\cdot)$ trained with the corresponding clean dataset are close to $1$, indicating our solution can give perfect detection with ideal features.

\subsection{More Experiments on Clothing1M}
Besides, we test the performance of training only with the clean instances selected by our approach in Table~\ref{table:c1m}. Standard training with Cross-Entropy loss is adopted. The only difference between the first row and other rows of Table~\ref{table:c1m} is that some training instances are filtered out by our approach. Table~\ref{table:c1m} shows simply filtering out corrupted instances based on our approach distinctively outperforms the baseline. We also observe that slightly tuning $g(\cdot)$ in the fine-grained Clothing1M dataset would be helpful.
Note the best-epoch test accuracy we can achieve is $73.64\%$, which outperforms many baselines such as HOC $73.39\%$ \citep{zhu2021clusterability}, GCE+SimCLR $73.35\%$ \citep{ghosh2021contrastive}, CORES $73.24\%$ \citep{cheng2021learningsieve}, GCE $69.75\%$ \citep{zhang2018generalized}.
See more detailed settings and discussions in Appendix~\ref{supp:c1m}.

%% file: src_icml/conclusion.tex
\section{Conclusions}
This paper proposed a new and universally applicable data-centric training-free solution to detect noisy labels by using the neighborhood information of features.
We have also demonstrated that the proposed data-centric method works even better than the learning-centric method when both methods are build on the same features.
Future works will explore other tasks that could benefit from label cleaning, e.g., fairness \cite{liu2021can} and multi-label learning \cite{liu2021emerging}, and extend the idea to long-tail sub-population detection \cite{liu2021importance,wei2022open}.

\paragraph{Acknowledgment}
This work is partially supported by the National Science Foundation (NSF) under grants IIS-2007951, IIS-2143895, and the Office of Naval Research under grant N00014-20-1-22.

%% file: src_icml/appendix.tex
\appendix

The omitted proofs and experiment settings are provided as follows.
\section{Theoretical Analyses}

\subsection{Proof for Proposition~\ref{thm:vote}}\label{proof:vote}
Now we derive a lower bound for the probability of getting true detection with majority vote:
\begin{equation*}
    \begin{split}
        \PP(\text{Vote is correct}|k) 
        \ge & (1-\delta_k) \cdot \bigg[ p \sum_{l=0}^{\lceil (k+1)/2 \rceil -1} {k+1 \choose l} e^l (1-e)^{k+1-l} \\
        & \quad + (1-p) \sum_{l=0}^{\lceil (k+1)/2 \rceil -1} {k+1 \choose l}  e^l (1-e)^{k+1-l}  \bigg] \\
        = &(1-\delta_k) \cdot \left[  p \cdot I_{1-e}(k+1-k', k'+1) 
        + (1-p) \cdot I_{1-e}(k+1-k', k'+1) \right]
    \end{split}
\end{equation*}
where $I_{1-e}(k+1-k',k'+1)$ is the regularized incomplete beta function defined as
\[
I_{1-e}(k+1-k',k'+1) = (k+1-k'){k+1\choose k'} \int_0^{1-e} t^{k-k'}(1-t)^{k'} dt,
\]
and
$k'=\lceil (k+1)/2 \rceil -1$.

\section{Proof for Theorem~\ref{thm:rank}}

\begin{proof}

Now we derive the worst-case error bound. We first repeat the notations defined in Section~\ref{sec:thm1} as follows.

Denote random variable $S$ by the score of each instance being clean. A higher score $S$ indicates the instance is more likely to be clean.
\rev{
Denote the score of a true/false instance by 
\begin{align*}
    & S^{\text{true}}_{n,j}:=\textsf{Score}(\hat \vy_n,j)~|~(\tilde y_n = y_n = j), \\
    & S^{\text{false}}_{n',j}:=\textsf{Score}(\hat \vy_{n'},j)~|~(\tilde y_{n'} = j, y_{n'} \ne j).
\end{align*}
Both are scalars.
Then for instances in $\mathcal N_j$, we have two set of random variables $\mathbb S_j^{\text{true}}:=\{S^{\text{true}}_{n,j}|n\in\mathcal N_j, \tilde y_n = y_n = j\}$ and $\mathbb S_j^{\text{false}}:=\{S^{\text{false}}_{n',j}|n'\in\mathcal N_j,\tilde y_{n'} =j, y_{n'} \ne j\}$.
Recall $\mathcal N_j :=\{n|\tilde y_n = j\}$ are the set of indices that correspond to noisy class $j$.
Intuitively, the score $S_{n,j}^{\text{true}}$ should be greater than $S_{n',j}^{\text{false}}$.
Suppose their means, which depend on noise rates, are bounded, i.e., $$\mathbb E[S_{n,j}^{\text{true}}] \ge \mu_j^{\text{true}}, \quad \mathbb E[S_{n',j}^{\text{false}}] \le \mu_j^{\text{false}}$$ for all feasible $n, n'$.
Assume there exists a feasible $v$ such that
both $S_j^{\text{true}}$  and $S_j^{\text{false}}$ follow sub-Gaussian distributions with variance proxy $\frac{\Delta^2}{2v}$ \citep{buldygin1980sub,zhu2021federated} such that:
$$\PP( \mu_j^{\text{true}} - S_{n,j}^{\text{true}} \ge t) \le e^{-\frac{vt^2}{\Delta^2}}, \PP(S_{n',j}^{\text{false}} - \mu_j^{\text{false}} \ge  t) \le e^{-\frac{vt^2}{\Delta^2}},$$ and the probability density satisfies $\PP(S_j^{\text{true}} = \mu_j^{\text{true}} ) = \PP(S_j^{\text{false}} = \mu_j^{\text{false}} ) = 1/\Delta$, where $1/\Delta$ is the ``height'' of both distributions, $v$ is the decay rate of tails.
Let $N_j^-$ ($N_j^+$) be the number of indices in $\mathbb S_j^{\text{false}}$ ($\mathbb S_j^{\text{true}}$).}

For ease of notations, we omit the subscript $j$ in this proof since the detection is performed on each $j$ individually.

Let $S^{\text{false}}$  ($S^{\text{true}}$) be an arbitrary random variable in $\mathbb S^{\text{false}}$ ($\mathbb S^{\text{true}}$).
Denote the order statistics of random variables in set $\mathbb S^{\text{false}}$ by $S^{\text{false}}_{(1)},\cdots S^{\text{false}}_{(N^-)}$, where $S^{\text{false}}_{(1)}$ is the smallest order statistic and $S^{\text{false}}_{(N^-)}$ is the largest order statistic.
The following lemma motivates the performance of the rank-based method. 
\begin{lem}\label{lem:fscore}
The $F_1$-score of detecting corrupted labels in $\mathcal N_j$ by the rank-based method will be no less than $1-\alpha/N^-$ when the true probability $\PP(Y=j|\widetilde Y=j)$ is known and 
$S^{\text{false}}_{(N^-)} < S^{\text{true}}_{(\alpha+1)}.$
\end{lem}
Lemma~\ref{lem:fscore} connects the upper bound for the number of wrongly detected corrupted instances with order statistics.
There are two cases that can cause detection errors:\\
{\bf Case-1}: 
\[
0\le \mu^{\text{true}} - S^{\text{true}} < \Delta \text{~~and~~} 0\le S^{\text{false}} - \mu^{\text{false}} < \Delta: \text{at most}~\alpha~  \text{errors when} ~ S^{\text{false}}_{(N^-)} < S^{\text{true}}_{(\alpha+1)}.
\]
and {\bf Case-2}:
\[
 \mu^{\text{true}} - S^{\text{true}} \ge \Delta \text{~~or~~} S^{\text{false}} - \mu^{\text{false}} \ge \Delta: \text{at most}~ \max(N_-,N_+) ~  \text{errors}
\]
We analyze each case as follows.

\paragraph{Case-1:}
When Case-1 holds, we have 
\[
\PP(  \mu^{\text{true}} - S^{\text{true}} = x ) \le 1/\Delta, x\in[0,\Delta]
\]
and
\[
\PP(S^{\text{false}} - \mu^{\text{true}} = x ) \le 1/\Delta, x\in[0,\Delta].
\]
The above two inequalities show that the left tail of $S^{\text{true}}$ and the right tail of $S^{\text{false}}$ can be upper bounded by uniform distributions. Denote the corresponding uniform distribution by $U^{\text{true}}\sim \textsf{Unif}(\mu^{\text{true}}-\Delta,\mu^{\text{true}})$ and $U^{\text{false}}\sim \textsf{Unif}(\mu^{\text{false}},\mu^{\text{false}}+\Delta)$. 

With true $\PP(Y=j|\widetilde Y=j)$, the detection errors only exist in the cases when the left tail of $S^{\text{true}}$ and the right tail of $S^{\text{false}}$ are overlapped. 
When the tails are upper bounded by uniform distributions, we have
\begin{equation*}
    \begin{split}
        & \PP(S^{\text{false}}_{(N^-)} < S^{\text{true}}_{(\alpha+1)})  \ge \PP(U^{\text{false}}_{(N^-)} < U^{\text{true}}_{(\alpha+1)})   \\
      = & \PP\left( 
        \left[U^{\text{false}} - \mu^{\text{false}}\right]_{(N^-)} + \mu^{\text{false}} < \left[U^{\text{true}} - (\mu^{\text{true}}-\Delta)\right]_{(\alpha+1)} + (\mu^{\text{true}}-\Delta)  \right)  \\
      = & \PP\left( 
        \left[U^{\text{false}} - \mu^{\text{false}}\right]_{(N^-)} -\left[U^{\text{true}} - (\mu^{\text{true}}-\Delta)\right]_{(\alpha+1)} <   \mu^{\text{true}} - \mu^{\text{false}} -\Delta  \right) .
    \end{split}
\end{equation*}
Note
\[
\left[U^{\text{false}} - \mu^{\text{false}}\right]_{(N^-)} \sim Beta(N^-,1),
\]
and
\[
\left[U^{\text{true}} - (\mu^{\text{true}}-\Delta)\right]_{(\alpha+1)}  \sim Beta(\alpha+1,N^+-\alpha),
\]
where $Beta$ denotes the Beta distribution.
Both variables are independent.
Thus the PDF of the difference is 
{\scriptsize
\begin{equation*}
    f(p) = 
    \begin{cases}
    B(N^+-\alpha,1)p^{N^+-\alpha}(1-p)^{\alpha+1} F(1,N^-+N^+,1-N^-;\alpha+2;1-p,1-p^2)/A, & 0 < p \le 1 \\
    B(N^--,N^+-\alpha)(-p)^{N^+-\alpha}(1+p)^{N^-+N^+-\alpha-1} F(N^+-\alpha,-\alpha,N^-+N^+;N^-+N^+-\alpha;1-p^2,1+p)/A, & -1 \le  p < 0  \\
    B(N^-+\alpha,N^+-\alpha)/A, & p=0,
    \end{cases}
\end{equation*}}
where $A=B(N^-,1)B(\alpha+1,N^+-\alpha)$, $B(a,b) = \int_{0}^1 t^{a-1}(1-t)^{b-1} dt$
\[
F(a,b_1,b_2;c; x,y) = \frac{\Gamma(c)} {\Gamma(a)\Gamma(c-a)} 
\int_0^1 t^{a-1} (1-t)^{c-a-1} (1-xt)^{-b_1} (1-yt)^{-b_2} \,\mathrm{d}t.
\]
Therefore, we have
\[
\PP(S^{\text{false}}_{(N^-)} < S^{\text{true}}_{(\alpha+1)})  \ge \int_{-1}^{\mu^{\text{true}} - \mu^{\text{false}} -\Delta} f(p) dp.
\]

\paragraph{Case-2}

The other part, we have no more than $ e^{-{v}} \cdot \max( N^-, N^+ )$ corrupted instances that may have higher scores than one clean instance.

\paragraph{Wrap-up}
From the above analyses, we know, w.p. at least $\int_{-1}^{\mu^{\text{true}} - \mu^{\text{false}} -\Delta} f(p) dp$, there are at most $e^{-{v}} \max( N^-, N^+ ) + \alpha$ errors in detection corrupted instances.
Note ${\sf Precision} = {\sf Recall}$ if we detect with the best threshold $N_j\PP(Y=j|\widetilde Y=j)$. 
Therefore, the corresponding $F_1$-score would be at least $1-\frac{e^{-{v}} \max( N^-, N^+ ) + \alpha}{N^-}$.

\end{proof}

\section{Experiment Settings on Clothing1M}\label{supp:c1m}
We firstly perform noise detection on 1 million noisy training instances then train only with the selected clean data to check the effectiveness.
Particularly, in each epoch of the noisy detection, we use a batch size of 32 and sample 1,000 mini-batches from 1M training instances while ensuring the (noisy) labels are balanced. We repeat noisy detection for $600$ epochs to ensure a full coverage of 1 million training instances. Parameter $k$ is set to 10.

\paragraph{Feature Extractor:}
We tested three different feature extractors in Table~\ref{table:c1m}: R50-Img, ViT-B/32-CLIP, and R50-Img Warmup-1. The former two feature extractors are the same as the ones used in Table~\ref{table:diff_pretrain}. Particularly, R50-Img means the feature extractor is the standard ResNet50 encoder (removing the last linear layer) pre-trained on ImageNet \citep{imagenet_cvpr09}. ViT-B/32-CLIP indicates the feature extractor is a vision transformer pre-trained by CLIP \citep{radford2021learning}. Noting that Clothing1M is a fine-grained dataset. To get better domain-specific fine-grained visual features, we slightly train the ResNet50 pre-trained with ImageNet for one epoch, i.e., 1,000 mini-batches (batch size 32) randomly sampled from 1M training instances while ensuring the (noisy) labels are balanced. The learning rate is 0.002.

\paragraph{Training with the selected clean instances:}
Given the selected clean instances from our approach, we directly apply the Cross-Entropy loss to train a ResNet50 initialized by standard ImageNet pre-trained parameters. We {\bf did not} apply any sophisticated training techniques, e.g., mixup \citep{zhang2018mixup}, dual networks \citep{Li2020DivideMix,han2018co}, loss-correction \citep{liu2015classification,natarajan2013learning,patrini2017making}, and robust loss functions \citep{liu2019peer,cheng2021learningsieve,zhu2020second,wei2020optimizing}. We train the model for $80$ epochs with a batch size of $32$. We sample $1,000$ mini-batches per epoch randomly selected from 1M training instances. Note Table~\ref{table:c1m} does not apply balanced sampling. Only the pure cross-entropy loss is applied. We also test the performance with balanced training, i.e., in each epoch, ensure the noisy labels from each class are balanced. Our approach can be consistently benefited by balanced training, and achieves an accuracy of $73.97$ in the best epoch, outperforming many baselines such as HOC $73.39\%$ \citep{zhu2021clusterability}, GCE+SimCLR $73.35\%$ \citep{ghosh2021contrastive}, CORES $73.24\%$ \citep{cheng2021learningsieve}, GCE $69.75\%$ \citep{zhang2018generalized}. We believe the performance could be further improved by using some sophisticated training techniques mentioned above.

\begin{table}[!t]
\caption{Experiments on Clothing1M \citep{xiao2015learning} {\bf with or without balanced sampling}. None: Standard training with 1M noisy data. R50-Img (or ViT-B/32-CLIP, R50-Img Warmup-1): Apply our method with ResNet50 pre-trained on ImageNet (or ViT-B/32 pre-trained by CLIP, R50-Img with 1-epoch warmup).}
\label{table:c1m_balanced}
\vskip 0.15in %
\begin{center}\vspace{-0pt}
\begin{small}
\begin{sc}
\scalebox{0.95}{\footnotesize
\begin{tabular}{c|cccc}
\hline
Data Selection             & \# Training Samples & Best Epoch & Last 10 Epochs                    & Last Epoch \\ \hline\hline
None (Standard Baseline) (Unbalanced)                       & 1M (100\%)          & 70.32      & 69.44 $\pm$ 0.13 & 69.53      \\ 
None (Standard Baseline) \quad (Balanced)                       & 1M (100\%)          & 72.20      & 71.40 $\pm$ 0.31 & 71.22      \\ 
R50-Img    (Unbalanced)      & 770k (77.0\%)       & 72.37      & 71.95 $\pm$ 0.08 & 71.89      \\ 
R50-Img \quad  (Balanced)          & 770k (77.0\%)       & 72.42      & 72.06  $\pm$ 0.16 & 72.24     \\ 
ViT-B/32-CLIP  (Unbalanced)            & 700k (70.0\%)       & 72.54      & 72.23 $\pm$ 0.17 & 72.11      \\ 
ViT-B/32-CLIP \quad  (Balanced)             & 700k (70.0\%)       & 72.99      & 72.76 $\pm$ 0.15& 72.91      \\ 
R50-Img Warmup-1 (Unbalanced)& 767k (76.7\%)       & {\bf 73.64}      & {\bf 73.28 $\pm$ 0.18} & {\bf 73.41}      \\
R50-Img Warmup-1 \quad (Balanced) & 767k (76.7\%)       & {\bf 73.97}      & {\bf 73.37  $\pm$ 0.03} & {\bf 73.35}      \\ \hline\hline
\end{tabular}}
\end{sc}
\end{small}
\end{center}
\vskip -0.1in %
\end{table}